\documentclass[sigconf]{acmart}

\copyrightyear{2026}
\acmYear{2026}
\setcopyright{cc}
\setcctype{by}
\acmConference[MM '26]{Proceedings of the 34th ACM International Conference on Multimedia}{November 10--14, 2026}{Rio de Janeiro, Brazil}
\acmBooktitle{Proceedings of the 34th ACM International Conference on Multimedia (MM '26), November 10--14, 2026, Rio de Janeiro, Brazil}
\acmDOI{10.1145/3767308.3836116}
\acmISBN{979-8-4007-2213-4/2026/11}

\usepackage{amsmath}
\usepackage{amssymb}
\usepackage{mathtools}

\usepackage{subcaption}
\usepackage{xspace}
\usepackage{xcolor}

\definecolor{weikaiGreen}{HTML}{196F3D}

\newcommand{\Sect}[1]{Section~\ref{#1}}
\newcommand{\Fig}[1]{Fig.~\ref{#1}}
\newcommand{\Tbl}[1]{Tbl.~\ref{#1}}
\newcommand{\Eqn}[1]{Eqn.~\ref{#1}}

\makeatletter
\DeclareRobustCommand\onedot{\futurelet\@let@token\@onedot}
\def\@onedot{\ifx\@let@token.\else.\null\fi\xspace}
\def\eg{\emph{e.g}\onedot}
\def\ie{\emph{i.e}\onedot}

\makeatother

\newcommand{\proj}{\textsc{UniMod}\xspace}

\newcommand{\Lcross}{\mathcal{L}_{\text{cross}}}
\newcommand{\Lwithin}{\mathcal{L}_{\text{within}}}
\newcommand{\Lclassify}{\mathcal{L}_{\text{cls}}}

\newcommand{\Ltotal}{\mathcal{L}_{\text{total}}}

\newcommand{\fimg}{f_{\text{img}}}
\newcommand{\ftxt}{f_{\text{txt}}}
\newcommand{\fmm}{f_{\text{mm}}}
\newcommand{\zimg}{\mathbf{z}_{\text{img}}}
\newcommand{\ztxt}{\mathbf{z}_{\text{txt}}}

\newcommand{\no}[1]{}
\newcommand{\RNum}[1]{\uppercase\expandafter{\romannumeral #1\relax}}

\begin{document}

\title{UniMod: Enhancing Multi-Modal Medical Diagnosis through Cross-Modality and Within-Modality Alignment}

\author{Zijian Gu}
\orcid{0009-0007-9591-7119}
\affiliation{%
  \institution{University of Central Florida}
  \city{Orlando}
  \state{FL}
  \country{United States}}
\email{zi700688@ucf.edu}

\author{Weikai Lin}
\orcid{0000-0003-3537-4857}
\affiliation{%
  \institution{University of Rochester}
  \city{Rochester}
  \state{NY}
  \country{United States}}
\email{wlin33@ur.rochester.edu}

\author{Shuang Zhou}
\orcid{0000-0001-5739-1637}
\affiliation{%
  \institution{Harvard Medical School}
  \city{Boston}
  \state{MA}
  \country{United States}}
\email{szhou18@mgh.harvard.edu}

\author{Zihan Chen}
\orcid{0009-0006-2899-9268}
\affiliation{%
  \institution{University of Virginia}
  \city{Charlottesville}
  \state{VA}
  \country{United States}}
\email{brf3rx@virginia.edu}

\author{Song Wang$^{\dagger}$}
\thanks{$^{\dagger}$Corresponding author.}
\orcid{0000-0003-1273-7694}
\affiliation{%
  \institution{University of Central Florida}
  \city{Orlando}
  \state{FL}
  \country{United States}}
\email{song.wang@ucf.edu}

\begin{abstract}
Multi-modal learning combining medical images and clinical text is promising for disease diagnosis.
However, standard multi-modal training leads to shortcut learning: models exploit the easier modality (\eg, diagnostic cues in text) while neglecting harder-to-learn features (\eg, subtle visual patterns).
We propose \proj, a framework that mitigates shortcut learning by requiring each modality to predict the diagnosis on its own.
It supervises image-only, text-only, and multi-modal classification simultaneously, so each modality must extract diagnostic features.
We add cross-modality alignment for knowledge transfer and within-modality supervised contrastive alignment over same-diagnosis patients.
On Harvard-Glaucoma, \proj reaches 0.850 AUC, outperforming OGM-GE and Gradient Blending by 1.6--1.8\%; on CheXpert Plus, it reaches 0.966 AUC, surpassing them by over 5\%.
\proj also extends to 5-class multi-label diagnosis without architectural change, improving mean AUC by 0.097 over CGGM.
\end{abstract}



\renewcommand{\shortauthors}{Gu et al.}

\maketitle

\section{Introduction}
\label{sec:intro}

\begin{figure}[!t]
    \centering
    \includegraphics[width=\columnwidth]{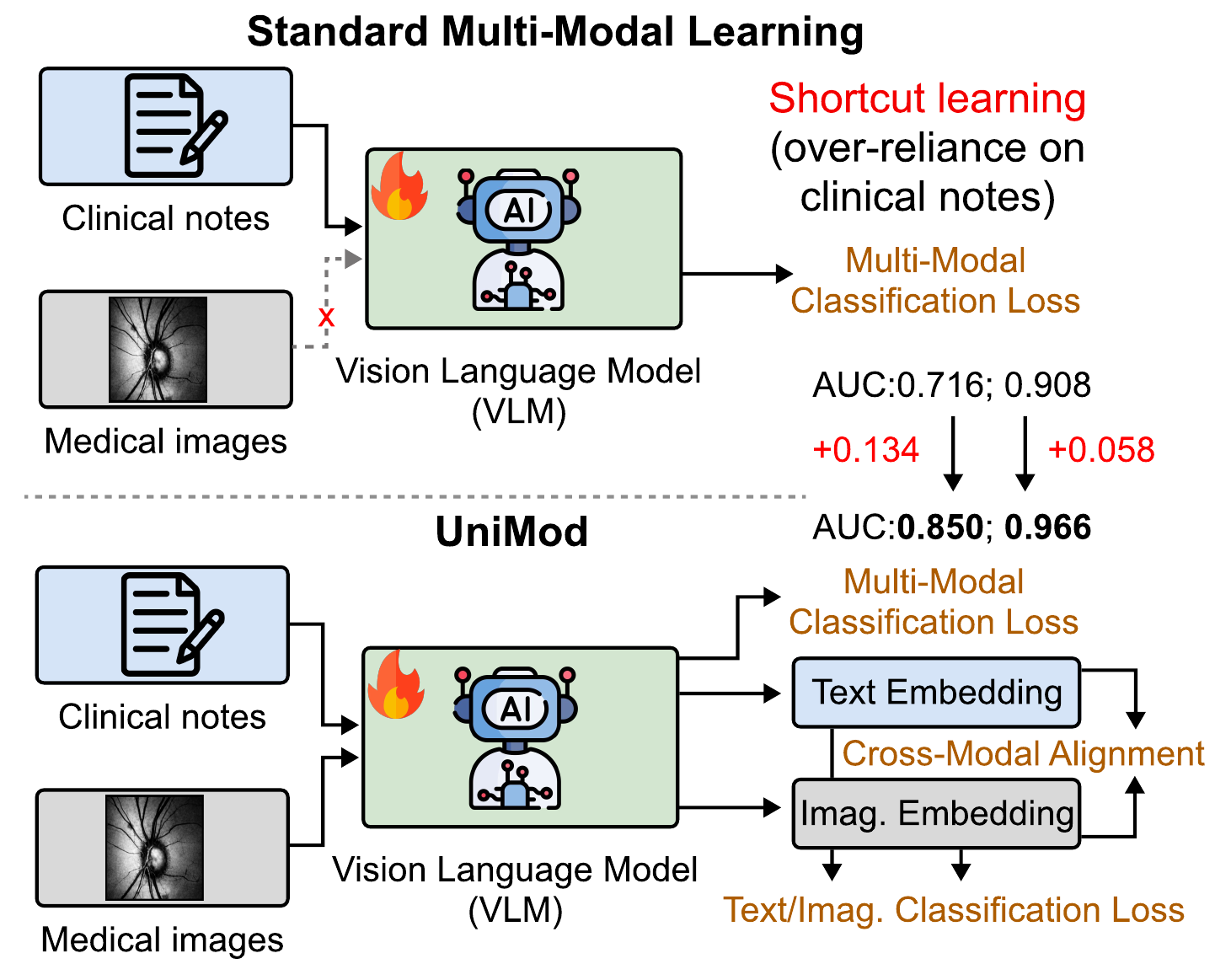}
    \caption{
        Comparison of Standard Multimodal LoRA and \proj.
        \textbf{Top:} Standard multi-modal fine-tuning takes shortcuts by relying primarily on text while ignoring visual features.
        \textbf{Bottom:} \proj extracts separate text and image embeddings, applies cross-modal alignment, and enforces unimodal classification losses, forcing each modality to learn meaningful diagnostic features independently. Fundus images \textcopyright{} Harvard Ophthalmology AI Lab.
    }
    \Description{Two-row schematic contrasting two fine-tuning strategies for a medical vision-language model. The top row shows standard multi-modal LoRA fine-tuning, in which a chest image and its clinical text are fused and a single multi-modal classifier is trained; the prediction path is dominated by the text branch while the image branch contributes little, illustrating shortcut learning. The bottom row shows UniMod, in which the image and text streams are kept separate to produce one embedding per modality, a cross-modality alignment objective links the two embeddings, and independent image-only, text-only, and multi-modal classification losses are applied so that each modality must predict the diagnosis on its own.}
    \label{fig:teaser}
    \vspace{-1em}
\end{figure}

Medical diagnosis increasingly relies on integrating multiple information sources: medical images (\eg, fundus photographs, X-rays, CT scans) and clinical text (\eg, physician notes, diagnostic reports).
Each modality provides unique diagnostic value. For example, images capture visual biomarkers while clinical notes encode physician observations and patient history.
Combining these modalities has the potential to improve diagnostic accuracy beyond what either modality can achieve alone~\cite{baltruvsaitis2018multimodal}.
Vision-language models (VLMs)~\cite{radford2021learning,liu2024visual,alayrac2022flamingo} have emerged as a powerful paradigm for integrating these heterogeneous modalities, achieving promising results across various medical tasks, including radiology report generation~\cite{tiu2022expert}, disease classification~\cite{wang2022medclip,huang2021gloria,rajpurkar2017chexnet}, and clinical decision support~\cite{wu2023medklip,singhal2023large}.

However, directly training multi-modal models often leads to \emph{shortcut learning}~\cite{geirhos2020shortcut,degrave2021ai}, which means the model exploits whichever modality provides the easiest path to correct predictions, while neglecting the other.
This phenomenon is particularly prevalent in medical diagnosis, where clinical text frequently contains explicit diagnostic cues (\eg, ``suspected glaucoma,'' ``abnormal finding''), whereas images require learning subtle visual patterns (\eg, optic disc cupping ratio, retinal nerve fiber layer defects).
When both modalities are available, the model takes the textual shortcut and fails to develop visual diagnostic capabilities.
As a result, two serious consequences could exist.
First, the model does not fully exploit each modality's diagnostic potential, leaving valuable information underutilized.
Second, performance degrades sharply when text is ambiguous, incomplete, or unavailable, a common clinical scenario where physicians may lack time for detailed notes, or a preliminary diagnosis must rely on imaging alone.

Existing approaches to multi-modal learning have attempted to address modality imbalance through various mechanisms.
Gradient blending~\cite{wang2020makes} adjusts learning rates based on modality-specific overfitting.
OGM-GE~\cite{peng2022balanced} balances gradients across modalities to prevent dominance.
Modality dropout~\cite{neverova2015moddrop} randomly masks input modalities during training to encourage robustness.
However, these methods do not explicitly require each modality to learn meaningful diagnostic features~\cite{xu2023multimodal}. In contrast, they only adjust the learning dynamics without fundamentally changing the training objective.
As a result, the model may still find shortcuts that satisfy the training loss without developing genuine modality-specific capabilities.

In this work, we address this challenge with a simple insight: \emph{if a model can make accurate predictions using each modality independently, it must have learned genuine diagnostic features from that modality}.
Based on this insight, we propose \proj, a framework that supervises image-only, text-only, and multi-modal predictions simultaneously.
This harder objective reduces shortcut learning: the image encoder must capture visual diagnostic features because it cannot fall back on textual cues, and each modality develops self-sufficient diagnostic capability that also benefits the combined prediction.

\proj consists of four complementary components.
We apply classification losses to all three prediction paths (image-only, text-only, and multi-modal), forcing each modality to extract diagnostic features independently (\Sect{sec:method:independent}).
Modality-separated attention keeps the two token streams apart so that these unimodal predictions are well defined (\Sect{sec:preliminary:framework}).
Cross-modality alignment pulls image and text representations of the same patient closer, enabling knowledge transfer between modalities (\Sect{sec:method:cross}).
Within-modality alignment applies supervised contrastive learning~\cite{khosla2020supcon,oord2018representation} to cluster same-diagnosis samples (\Sect{sec:method:within}).
%
%
We evaluate \proj on two medical diagnosis benchmarks: glaucoma detection using fundus images and clinical notes from the Harvard Glaucoma dataset~\cite{luo2023harvard}, and pleural effusion classification using chest X-rays and radiology reports from CheXpert Plus~\cite{chambon2024chexpert}, improving on modality-balancing baselines in accuracy and F1 on both datasets.

\proj is assembled from established components---independent unimodal supervision, MSE alignment, supervised contrastive learning, GradNorm, and LoRA~\cite{khosla2020supcon,chen2018gradnorm,hu2022lora}. In summary, our contributions are as follows:
\begin{itemize}
    \item a problem formulation: \emph{modality shortcut learning} in vision-language models whose two streams share a single backbone, which is distinct from the gradient-level modality imbalance targeted by prior balancing methods~\cite{peng2022balanced,wang2020makes};
    \item a framework that removes the lowest-loss route to that shortcut by supervising the three prediction paths independently, together with the modality separation that makes those unimodal objectives well defined;
    \item evidence that gradient modulation does not substitute for independent supervision---replacing it under the same alignment losses reaches 0.849 AUC against 0.857 for the full model---and that \proj transfers to 5-class multi-label diagnosis without architectural change.
\end{itemize}

\section{Related Work}
\label{sec:related}

\paragraph{Multi-Modal Medical Learning.}
Vision-language models (VLMs)~\cite{wang2025internvl3, chen2024expanding, bai2025qwen2, comanici2025gemini, applin2023gpt} have achieved remarkable success in medical imaging by combining visual and textual information~\cite{bommasani2021opportunities}.
Large-scale VLMs such as LLaVA~\cite{liu2024visual} and Flamingo~\cite{alayrac2022flamingo}, built on powerful vision encoders~\cite{dosovitskiy2020image,liu2021swin} and large language models~\cite{touvron2023llama}, demonstrate strong visual reasoning, and medical adaptations including LLaVA-Med~\cite{li2024llava}, BiomedCLIP~\cite{zhang2023biomedclip}, and Med-PaLM~\cite{singhal2023large} further specialize these models for clinical tasks.
MedCLIP~\cite{wang2022medclip} addresses false negatives in medical image-text pairs through semantic matching, while GLoRIA~\cite{huang2021gloria} introduces attention-weighted contrastive learning for chest X-ray interpretation.
Others scale medical vision-language pre-training on biomedical literature and radiology corpora~\cite{lin2023pmc,you2023cxr} or learn joint representations by reconstruction~\cite{chen2022multi}, while general-purpose models~\cite{li2023blip,kim2021vilt} and fusion strategies ranging from early concatenation~\cite{kan2026pamf,kan2026trace,zheng2026mutebench} to attention-based mechanisms~\cite{lu2019vilbert,chen2020uniter,xu2023multimodal,yi2023deepsta,yi2024learning} explore efficient designs.
However, a persistent challenge is ``modality laziness''~\cite{liang2022mind}, where models over-rely on one modality while ignoring others.
Solutions such as gradient blending~\cite{wang2020makes} and OGM-GE~\cite{peng2022balanced} adjust learning dynamics, but do not guarantee each modality learns meaningful features.
Our approach differs by enforcing accurate single-modality predictions, ensuring genuine feature extraction from each modality.

\paragraph{Shortcut Learning in Medical Diagnosis.}
Shortcut learning~\cite{geirhos2020shortcut,degrave2021ai} occurs when models exploit superficial correlations rather than learning robust diagnostic features.
In multi-modal medical diagnosis, text often contains explicit diagnostic cues (\eg, ``suspected glaucoma''), providing easier paths to correct predictions than learning subtle visual patterns.
Modality dropout~\cite{neverova2015moddrop} encourages robustness to missing inputs but does not explicitly require strong single-modality performance.
Distributionally robust optimization~\cite{sagawa2020distributionally} addresses worst-case group performance but does not target modality-specific feature learning.
Cross-modal distillation~\cite{huo2024c2kd} transfers knowledge from modality-rich to modality-limited models using fixed teachers.
In contrast, we adopt a mutual learning paradigm~\cite{zhang2018deep}: all prediction paths (image-only, text-only, multi-modal) learn simultaneously without a fixed teacher, enabling bidirectional knowledge transfer.

\paragraph{Representation Alignment.}
Self-supervised methods~\cite{chen2020simple,he2020moco,grill2020bootstrap,zbontar2021barlow} learn representations by contrasting or aligning positive pairs, with alignment and uniformity identified as the key properties~\cite{wang2020understanding}.
In the vision-language domain, CLIP~\cite{radford2021learning} and \citet{jia2021scaling} align image-text pairs in a shared embedding space at scale, and \citet{li2021align} propose aligning representations before fusion for improved multi-modal understanding.
The InfoNCE objective~\cite{oord2018representation} underlies these losses, and supervised contrastive learning~\cite{khosla2020supcon} extends them by treating same-class samples as positives.
We combine cross-modality alignment (same patient, different modalities) with within-modality alignment (same diagnosis, different patients), giving instance-level and class-level supervision of the representation space.


\begin{figure*}[!t]
    \centering
    \includegraphics[width=0.95\textwidth]{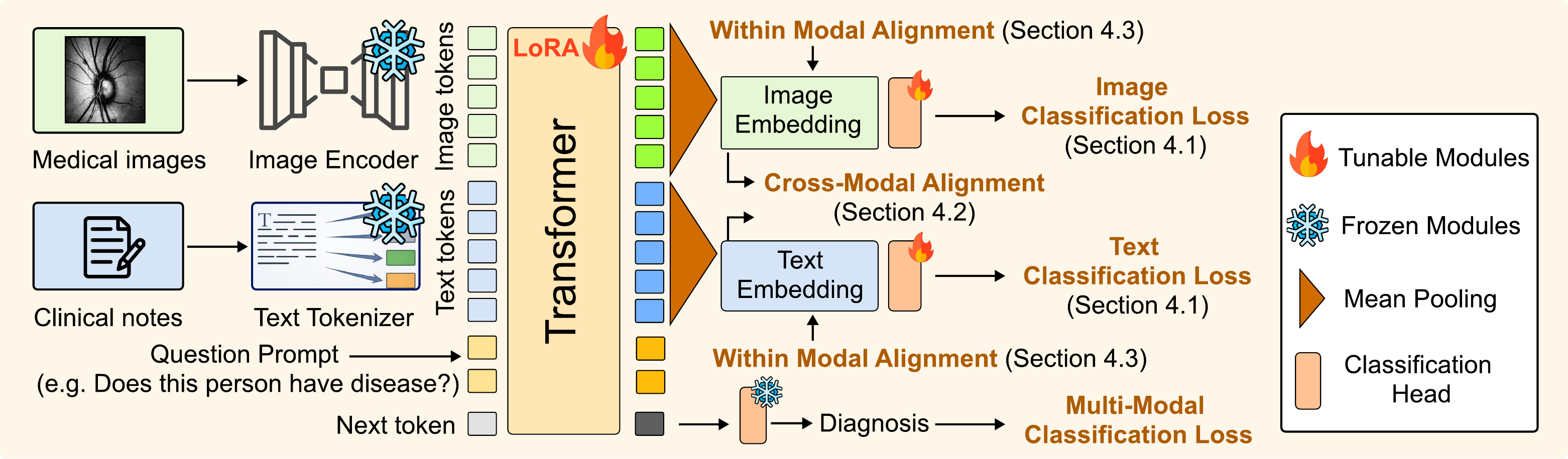}
    \caption{
        \textbf{Overview of \proj.}
        We implement a custom attention mask to enforce  modality isolation, yielding a unimodal embedding for each modality.
        We apply three losses: (1)~independent classification losses on image-only, text-only, and multi-modal predictions; (2)~cross-modality alignment ($\Lcross$) for cross-modality knowledge transfer; and (3)~within-modality contrastive alignment ($\Lwithin$) for class-level clustering. Fundus images \textcopyright{} Harvard Ophthalmology AI Lab.
    }
    \Description{Architecture diagram of the UniMod framework, read left to right. A medical image and its clinical text are tokenized and passed through a vision-language backbone whose self-attention uses a custom modality-isolation mask, drawn as a block-diagonal grid, so that image tokens and text tokens cannot attend to each other. This yields one unimodal embedding per modality. The two embeddings feed three groups of objectives: independent classification heads producing image-only, text-only, and multi-modal predictions, each with its own cross-entropy loss; a cross-modality alignment loss L-cross that pulls the paired image and text embeddings together; and a within-modality contrastive loss L-within that clusters embeddings of samples sharing the same diagnosis.}
    \label{fig:framework}
\end{figure*}

\section{Preliminary}
\label{sec:preliminary}

\subsection{Problem Setup}
\label{sec:preliminary:problem}

Each patient sample pairs a medical image $x_{\text{img}} \in \mathcal{X}_{\text{img}}$ (\eg, chest X-ray or fundus photograph) with clinical text $x_{\text{txt}} \in \mathcal{X}_{\text{txt}}$ (\eg, clinical notes or diagnostic reports).
Given $\mathcal{D} = \{(x_{\text{img}}^i, x_{\text{txt}}^i, y^i)\}_{i=1}^{N}$, we learn $f: \mathcal{X}_{\text{img}} \times \mathcal{X}_{\text{txt}} \rightarrow [0,1]^C$ predicting the probability of each of $C$ diagnostic labels; $C=1$ recovers the binary case, and $C>1$ the multi-label setting of \Sect{sec:exp:main}.

\subsection{Multi-Modal Learning Framework}
\label{sec:preliminary:framework}

We adopt a vision-language model (VLM) architecture $\fmm$ consisting of three components:
(1) an image encoder $\fimg: \mathcal{X}_{\text{img}} \rightarrow \mathbb{R}^{L_v \times d}$ that extracts a sequence of $L_v$ visual tokens from medical images,
(2) a text tokenizer $\ftxt: \mathcal{X}_{\text{txt}} \rightarrow \mathbb{R}^{L_t \times d}$ that converts clinical text into $L_t$ text tokens, and
(3) an autoregressive transformer backbone~\cite{vaswani2017attention} that processes the concatenated token sequence and predicts the next token.
Here $d$ is the dimension size.

Specifically, given an image $x_{\text{img}}$ and clinical text $x_{\text{txt}}$, the image encoder produces visual tokens $\mathbf{V} = \fimg(x_{\text{img}}) \in \mathbb{R}^{L_v \times d}$, and the text tokenizer produces text tokens $\mathbf{T} = \ftxt(x_{\text{txt}}) \in \mathbb{R}^{L_t \times d}$.
For diagnostic prediction, we append a query prompt $x_q$ (\eg, ``Does this patient have the disease? Answer:'') tokenized as $\mathbf{Q} = \ftxt(x_q) \in \mathbb{R}^{L_q \times d}$, followed by a prediction placeholder token $\mathbf{p} \in \mathbb{R}^d$.
The VLM processes the concatenated sequence and produces contextualized representations:
\begin{equation}
    \mathbf{H} = \fmm([\mathbf{V}; \mathbf{T}; \mathbf{Q}; \mathbf{p}]) \in \mathbb{R}^{(L_v + L_t + L_q + 1) \times d},
    \label{eq:vlm_input}
\end{equation}
where $[\cdot;\cdot]$ denotes sequence concatenation.
The transformer backbone employs causal attention, where each token can only attend to preceding tokens.
Crucially, we design a custom attention mask to enforce modality isolation during representation learning.
Image tokens $\mathbf{V}$ can only attend to other image tokens, and text tokens $(\mathbf{T}, \mathbf{Q})$ can only attend to other text tokens. 
As a result, neither modality can directly access the other, and only the final prediction token $\mathbf{p}$ is permitted to attend to tokens from both modalities.
This design ensures that the output representations $\mathbf{H}_{1:L_v}$ and $\mathbf{H}_{L_v+1:L_v+L_t}$ encode purely unimodal information, forcing each modality to develop self-contained, modality-specific features.

To extract modality-specific representations, we employ mean pooling over the corresponding output tokens.
Let $\mathbf{H}_{1:L_v}$ denote the output tokens corresponding to visual inputs and $\mathbf{H}_{L_v+1:L_v+L_t}$ denote those corresponding to text inputs.
The image and text representations are computed as:
\begin{equation}
    \zimg = \frac{1}{L_v} \sum_{i=1}^{L_v} \mathbf{H}_i, \quad
    \ztxt = \frac{1}{L_t} \sum_{i=L_v+1}^{L_v+L_t} \mathbf{H}_i.
    \label{eq:representations}
\end{equation}
\paragraph{Unimodal prediction.}
For single-modality classification, the unimodal representations are passed through a classification head $h: \mathbb{R}^d \rightarrow [0,1]$, yielding predictions $\hat{y}_{\text{img}} = h(\zimg)$ and $\hat{y}_{\text{txt}} = h(\ztxt)$.

\paragraph{Multi-modal prediction.}
For the final multi-modal prediction, we leverage the VLM's autoregressive next-token prediction capability.
The output hidden state at the prediction position $\mathbf{h} = \mathbf{H}_{L_v + L_t + L_q + 1} \in \mathbb{R}^d$ aggregates information from all preceding tokens, with cross-modal information only accessible at this stage due to our attention mask design.
The VLM's language modeling head produces logits over the vocabulary, and the multi-modal prediction probability is derived from the ``Yes'' and ``No'' tokens:
\begin{equation}
    \hat{y}_{\text{mm}} = \frac{\exp(\ell_{\text{Yes}})}{\exp(\ell_{\text{Yes}}) + \exp(\ell_{\text{No}})},
    \label{eq:prediction}
\end{equation}
where $\ell_{\text{Yes}}$ and $\ell_{\text{No}}$ denote the logits for the ``Yes'' and ``No'' tokens, respectively.
This formulation directly utilizes the VLM's pre-trained language modeling head without introducing extra parameters.

\subsection{The Shortcut Learning Problem}
\label{sec:preliminary:problem_analysis}

Standard multi-modal training directly optimizes the classification loss using the fused multi-modal representation.
While this approach is straightforward, it often leads to \emph{shortcut learning}~\cite{geirhos2020shortcut}: the model exploits superficial correlations between modalities rather than learning deep, modality-specific features.

Consider a glaucoma detection task with fundus images and clinical notes.
A naive multi-modal model might learn to rely primarily on explicit disease mentions in the text (\eg, ``suspected glaucoma'') while ignoring subtle but important visual biomarkers in the image (\eg, optic disc cupping ratio).
When the text modality is ambiguous or unavailable, such a model fails catastrophically.
This shortcut learning problem manifests in two specific ways.

\paragraph{Lack of cross-modal knowledge transfer.}
When the model can achieve good performance using only the text modality, it has no incentive to learn from images.
The image encoder becomes ``lazy'' and fails to capture diagnostic visual patterns.
The model over-relies on text, performing well with both modalities but degrading sharply when only one is available.

\paragraph{Unstructured within-modality representations.}
Even when each modality contributes, the representations lack semantic organization.
The model encodes imaging conditions, camera angles, or writing styles rather than diagnostic features.
Samples with the same diagnosis are scattered in the representation space rather than clustered together, making the decision boundary fragile.

\section{Method}
\label{sec:method}

We present \proj, a multi-modal learning framework designed to address shortcut learning in medical diagnosis (\Sect{sec:preliminary:problem_analysis}).
Our key insight is that \emph{requiring each modality to make accurate predictions discourages the model from taking shortcuts}, thereby learning genuine diagnostic features from both modalities.

As shown in \Fig{fig:framework}, \proj applies classification losses to image-only, text-only, and multi-modal predictions separately (\Sect{sec:method:independent}), pulls image and text representations closer via cross-modality alignment, enabling knowledge transfer across modalities (\Sect{sec:method:cross}), and clusters same-class samples via within-modality contrastive learning to form a better embedding structure (\Sect{sec:method:within}).
We finally combine all objectives into a unified training loss, balanced via dynamic weight adjustment using GradNorm~\cite{chen2018gradnorm} (\Sect{sec:method:objective}), following multi-task learning principles~\cite{ruder2017overview,kendall2018multi}.
To preserve the pre-trained knowledge of the VLM while adapting it to the target domain, we fine-tune the model weights using low-rank adaptation (\Sect{sec:method:lora}).

\subsection{Independent Modality Feature Extraction}
\label{sec:method:independent}

As discussed in \Sect{sec:preliminary:problem_analysis}, standard multi-modal training suffers from shortcut learning: the model relies on whichever modality provides the easiest path to low training loss, leaving the other modality underutilized.
For instance, when text contains explicit diagnostic cues (\eg, ``suspected glaucoma''), the model may ignore subtle but important visual patterns in the fundus image.

Our key insight is to force each modality to independently extract features sufficient for accurate classification.
Rather than allowing the model to ``take shortcuts'' by relying on cross-modal information, we require that each modality alone can make correct predictions.
Specifically, we apply classification losses to both the image and text representations:
\begin{equation}
    \Lclassify^{\text{img}} = \mathcal{L}_{\text{CE}}(\hat{y}_{\text{img}}, y), \quad \Lclassify^{\text{txt}} = \mathcal{L}_{\text{CE}}(\hat{y}_{\text{txt}}, y),
    \label{eq:unimodal_loss}
\end{equation}
where $\hat{y}_{\text{img}} = h(\zimg)$ and $\hat{y}_{\text{txt}} = h(\ztxt)$ are the unimodal predictions from \Eqn{eq:representations}, and $\mathcal{L}_{\text{CE}}$ denotes the cross-entropy loss.
Additionally, the multi-modal prediction $\hat{y}_{\text{mm}}$ from \Eqn{eq:prediction} is supervised with:
\begin{equation}
    \Lclassify^{\text{mm}} = \mathcal{L}_{\text{CE}}(\hat{y}_{\text{mm}}, y).
    \label{eq:mm_loss}
\end{equation}

By supervising each modality independently, we remove the lowest-loss route to a shortcut solution.
The image encoder must learn to capture visual diagnostic features (\eg, optic disc cupping, retinal nerve fiber layer defects) because it cannot rely on textual cues.
Similarly, the text encoder must learn to extract diagnostic information from clinical notes without depending on visual evidence.
This harder objective yields stronger representations: each modality develops self-sufficient diagnostic capability.
In summary, our framework employs three classification losses that are combined as:
\begin{equation}
    \Lclassify = \Lclassify^{\text{mm}} + \Lclassify^{\text{img}} + \Lclassify^{\text{txt}}.
    \label{eq:classify_total}
\end{equation}

\paragraph{Why separate the streams, and why still a VLM?}
The backbone is causal over $[\mathbf{V};\mathbf{T};\mathbf{Q};\mathbf{p}]$, so image tokens never attend to text in either case; the mask of \Sect{sec:preliminary:framework} removes only the remaining direction, text attending to image.
That direction is exactly what the unimodal objectives depend on: without the mask $\ztxt$ is a text representation that has already read the image, and $\Lclassify^{\text{txt}}$ is no longer a text-only prediction.
Modality separation is therefore a correctness precondition rather than an accuracy device, and the ablation behaves accordingly.
Removing it shifts multi-modal AUC by $-0.0017$ on Harvard-Glaucoma and $+0.0003$ on CheXpert Plus, both inside the seed-to-seed standard deviation, and leaves text-only AUC essentially unchanged ($0.855\rightarrow0.855$ and $0.964\rightarrow0.965$): the text branch was not in fact exploiting the visual stream, and the mask is what licenses that statement rather than assuming it.
Fusion is relocated, not removed---cross-modal interaction persists at the prediction token $\mathbf{p}$, which attends to both streams, and through $\Lcross$, which aligns the two representation spaces directly.
This is also why the VLM backbone is retained: its value here is the cross-modal grounding held in the pre-trained weights, which LoRA adapts over $0.1\%$ of parameters and which places $\zimg$ and $\ztxt$ in one $d$-dimensional space---the property that lets $\Lcross$ operate without a learned projection.

\subsection{Cross-Modality Alignment}
\label{sec:method:cross}

While independent feature extraction ensures each modality captures diagnostic information, it does not facilitate knowledge transfer between modalities.
In multi-modal learning, knowledge from one modality can benefit another: textual descriptions can guide the image encoder to focus on relevant visual patterns, and visual evidence can help disambiguate vague clinical notes.
Without explicit encouragement, such knowledge transfer does not occur naturally.

To promote cross-modal knowledge transfer, we introduce an alignment loss that encourages the image and text representations to be similar for the same patient.
Since our VLM architecture produces representations of the same dimensionality $d$ for both modalities, we directly minimize their distance without projection layers:
\begin{equation}
    \Lcross = \frac{1}{|B|} \sum_{i \in B} \| \zimg^i - \ztxt^i \|_2^2,
    \label{eq:cross_loss}
\end{equation}
where $B$ is the mini-batch.
This loss pulls the image and text representations of the same patient closer, encouraging both modalities to encode similar diagnostic information in a shared semantic space.

Cross-modality alignment enables mutual knowledge transfer.
Consider an image showing subtle optic disc changes that are difficult to detect visually but are explicitly described in the clinical notes (\eg, ``increased cup-to-disc ratio'').
Through alignment, the image encoder learns to associate these visual patterns with the corresponding textual descriptions, effectively transferring diagnostic knowledge from text to image.
Conversely, when clinical notes are ambiguous, visual evidence can provide complementary information through the aligned representation space.

\subsection{Within-Modality Alignment}
\label{sec:method:within}

Beyond cross-modal alignment, we also structure the representation space within each modality.
Intuitively, two fundus images with similar glaucoma severity should have similar representations, regardless of imaging conditions or patient demographics.
Such a structure makes the decision boundary more robust and generalizes better.

To achieve this, we apply supervised contrastive learning~\cite{khosla2020supcon}, which pulls same-class samples together while pushing different-class samples apart.
For each modality $m \in \{\text{img}, \text{txt}\}$:
\begin{equation}
    \Lwithin^{m} = -\frac{1}{|B|} \sum_{i \in B} \frac{1}{|P(i)|} \sum_{p \in P(i)} \log \frac{\exp(s_{ip} / \tau)}{\sum_{j \neq i} \exp(s_{ij} / \tau)},
    \label{eq:supcon}
\end{equation}
where $s_{ij} = \text{sim}(\mathbf{z}_i^m, \mathbf{z}_j^m)$ denotes cosine similarity between samples $i$ and $j$, $P(i) = \{p \in B : y_i = y_p, p \neq i\}$ is the set of positive samples (same label) for anchor $i$, and $\tau$ is a temperature parameter.
The total within-modality loss is $\Lwithin = \Lwithin^{\text{img}} + \Lwithin^{\text{txt}}$.

This loss pulls samples with the same label together while pushing samples with different labels apart.
The resulting representation space exhibits clear semantic structure: healthy and diseased samples form distinct clusters.

Within-modality alignment is complementary to cross-modality alignment.
Cross-modality alignment operates at the sample level, aligning different views of the same patient.
Within-modality alignment operates at the class level, aligning different patients with the same diagnosis.
Together, they provide both instance-level and class-level supervision.

\subsection{Overall Training Objective}
\label{sec:method:objective}

We combine the classification and alignment losses into a unified training objective:
\begin{equation}
    \Ltotal = \Lclassify + \lambda_2 \Lcross + \lambda_3 \Lwithin,
    \label{eq:total_loss}
\end{equation}
where $\lambda_2, \lambda_3 > 0$ are weighting coefficients.
We fix $\lambda_1 = 1$ to use the classification loss as a stable reference scale, ensuring that the primary diagnostic objective maintains consistent importance throughout training.
Manually tuning $\lambda_2$ and $\lambda_3$ is challenging, as the optimal balance depends on the dataset and training dynamics~\cite{kendall2018multi}.
We adopt GradNorm~\cite{chen2018gradnorm} to adaptively adjust the relative weights $\lambda_2$ and $\lambda_3$ during training.
GradNorm monitors the gradient magnitudes of each loss component and adjusts the alignment loss weights to ensure balanced learning progress across all objectives.

\subsection{Parameter-Efficient Medical Adaptation}
\label{sec:method:lora}

Large-scale VLMs are pre-trained on general-domain data (\eg, natural images and web text) without exposure to medical images or clinical terminology~\cite{touvron2023llama,bommasani2021opportunities}.
Directly applying such models to medical diagnosis yields suboptimal performance due to the domain gap.
A straightforward solution is to fine-tune the entire VLM using the losses defined above (\Eqn{eq:total_loss}).
However, medical datasets are typically small, making full fine-tuning prone to catastrophic forgetting: the model loses its pre-trained knowledge, overfitting to limited medical samples and failing to generalize to new data.

To address this, we adopt Low-Rank Adaptation (LoRA)~\cite{hu2022lora}, a parameter-efficient fine-tuning method among alternatives such as adapters~\cite{houlsby2019parameter}, prefix tuning~\cite{li2021prefix}, and QLoRA~\cite{dettmers2023qlora}.
LoRA freezes the pre-trained weights $W_0$ and injects trainable low-rank matrices:
\begin{equation}
    W = W_0 + \Delta W = W_0 + BA,
    \label{eq:lora}
\end{equation}
where $B \in \mathbb{R}^{d \times r}$ and $A \in \mathbb{R}^{r \times k}$ are low-rank matrices with $r \ll \min(d, k)$.
This preserves pre-trained knowledge in $W_0$ while $BA$ learns domain-specific adaptations.

We apply LoRA to all projection matrices in the attention layers ($W_Q$, $W_K$, $W_V$, and $W_O$) while keeping the feed-forward network (FFN) layers frozen.
This design allows the model to fully adapt its attention patterns to medical-specific features while preserving the general-purpose transformations learned during pre-training.


\begin{table*}[!t]
    \centering
    \caption{Main results on Harvard-Glaucoma and CheXpert Plus datasets. We report AUC, Accuracy, F1, Precision, and Recall. Best results in \textbf{bold}.}
    \vspace{-1em}
    \label{tbl:main_results}
    \begin{tabular}{l|ccccc|ccccc}
    \toprule
    & \multicolumn{5}{c|}{\textbf{Harvard-Glaucoma}} & \multicolumn{5}{c}{\textbf{CheXpert Plus}} \\
    \textbf{Method} & AUC & ACC & F1 & Prec. & Rec. & AUC & ACC & F1 & Prec. & Rec. \\
    \midrule
    Zero-shot & 0.472 & 0.512 & 0.677 & 0.512 & \textbf{1.000} & 0.561 & 0.501 & 0.667 & 0.501 & \textbf{0.999} \\
    Image-only & 0.734 & 0.669 & 0.709 & 0.644 & 0.788 & 0.909 & 0.842 & 0.842 & 0.840 & 0.845 \\
    Text-only & 0.814 & 0.724 & 0.766 & 0.676 & 0.884 & 0.961 & 0.908 & 0.909 & 0.891 & 0.928 \\
    Multimodal LoRA & 0.716 & 0.640 & 0.720 & 0.598 & 0.904 & 0.908 & 0.861 & 0.862 & 0.855 & 0.869 \\
    OGM-GE~\cite{peng2022balanced} & 0.835 & 0.752 & 0.728 & 0.828 & 0.649 & 0.918 & 0.847 & 0.851 & 0.829 & 0.875 \\
    G-Blend~\cite{wang2020makes} & 0.837 & 0.732 & 0.686 & \textbf{0.857} & 0.572 & 0.919 & 0.842 & 0.854 & 0.793 & 0.926 \\
    \midrule
    \proj (Ours) & \textbf{0.850} & \textbf{0.768} & \textbf{0.786} & 0.745 & 0.831 & \textbf{0.966} & \textbf{0.919} & \textbf{0.920} & \textbf{0.909} & 0.931 \\
    \bottomrule
    \end{tabular}
    \vspace{-1em}
    \end{table*}
    
\begin{table}[!t]
    \caption{Dataset statistics for the binary setting; all inputs are paired image--text. CheXpert Plus is additionally evaluated as a 5-class multi-label task (\Tbl{tbl:multilabel}).}
    \label{tbl:datasets}
    \vspace{-1em}
    \centering
    \begin{tabular}{lcccc}
        \toprule
        Dataset & Train & Val & Test & Pos\% \\
        \midrule
        Harvard-Glaucoma & 7,000 & 1,000 & 2,000 & 50.5\% \\
        CheXpert Plus & 7,000 & 1,000 & 2,000 & 50.0\% \\
        \bottomrule
    \end{tabular}
\end{table}

\section{Experiments}
\label{sec:experiments}

We evaluate \proj on two medical diagnosis benchmarks: glaucoma detection from fundus images and pleural effusion detection from chest X-rays.
We first describe the experimental setup in \Sect{sec:exp:setup}, then demonstrate that \proj consistently outperforms baselines on both datasets (\Sect{sec:exp:main}).
We then measure modality reliance directly and quantify how far \proj reduces the text shortcut (\Sect{sec:exp:shortcut}), conduct ablation studies to analyze each component's contribution (\Sect{sec:exp:ablation}), and present qualitative case studies (\Sect{sec:exp:case}).

\subsection{Experimental Setup}
\label{sec:exp:setup}

\paragraph{Datasets.}
We use two datasets with paired medical images and clinical text.
\textbf{Harvard-Glaucoma}~\cite{luo2023harvard} contains fundus photographs paired with clinical notes for glaucoma detection, a globally prevalent eye disease~\cite{tham2014global}.
\textbf{CheXpert Plus}~\cite{chambon2024chexpert} contains chest X-ray images paired with radiology reports for pleural effusion detection, building on the CheXpert benchmark~\cite{irvin2019chexpert}.
Both provide binary labels; we additionally evaluate CheXpert Plus as a 5-class multi-label task (\Sect{sec:exp:main}).
\Tbl{tbl:datasets} summarizes the statistics.

\paragraph{Addressing potential label leakage.}
In both datasets, clinical text may contain explicit diagnostic cues that could leak label information.
For Harvard-Glaucoma, 98.6\% of clinical notes contain the word ``glaucoma,'' and 15.2\% contain statements like ``has glaucoma'' despite a negative label (due to historical records).
For CheXpert Plus, 97\% of radiology impressions mention ``effusion.''
To prevent such leakage, we implement text cleaning that removes diagnosis-related keywords, medication names, and procedure terms from clinical text before training.
Specifically, for Harvard-Glaucoma, we remove glaucoma-related terms, 14 diagnosis phrases, 14 medication patterns (\eg, latanoprost, timolol), and 11 procedure patterns (\eg, trabeculectomy, iridotomy).
For CheXpert Plus, we filter out sentences containing ``effusion'' or ``pleural fluid.''
After cleaning, 0\% of samples contain direct label leakage, and all experiments use cleaned text by default.

\paragraph{Baselines.}
We compare against six baselines.
\textbf{Zero-shot} directly uses the pre-trained VLM without fine-tuning.
\textbf{Image-only} and \textbf{Text-only} fine-tune with a single modality.
\textbf{Multimodal LoRA} fine-tunes with standard LoRA~\cite{hu2022lora} using both modalities.
\textbf{OGM-GE}~\cite{peng2022balanced} applies on-the-fly gradient modulation to suppress the dominant modality based on prediction confidence ratios.
\textbf{G-Blend}~\cite{wang2020makes} estimates optimal loss weights based on per-modality overfitting-to-generalization ratios.
All methods use the same VLM backbone and LoRA configuration for fair comparison.

\paragraph{Implementation.}
We fine-tune InternVL2.5-8B~\cite{chen2024expanding} (256 visual tokens per image) for 3 epochs with AdamW, learning rate $5\times10^{-5}$, effective batch size 32, LoRA~\cite{hu2022lora} rank $r=32$ and $\alpha=64$, SupCon temperature $\tau=0.07$, and GradNorm-adjusted loss weights~\cite{chen2018gradnorm}, on 2 NVIDIA RTX 5090 GPUs.

\paragraph{Metrics.}
We report AUC (area under the ROC curve)~\cite{fawcett2006roc}, F1 score, and accuracy.
AUC is the primary metric as it is threshold-independent and robust to class imbalance~\cite{he2009imbalanced}.

\begin{table*}[!t]
    \centering
    \begin{minipage}[t]{0.62\textwidth}
        \caption{Ablation study on \proj components. Each row removes only one component to  evaluate each component's contribution.}
        \label{tbl:ablation}
        
    \end{minipage}%
    \hfill
    \begin{minipage}[t]{0.35\textwidth}
        \caption{Robustness to missing modalities (AUC). $\Delta$: our improvement over MM LoRA.}
        \label{tbl:robustness}
    \end{minipage}
    \vspace{0.5em}
    \begin{minipage}[c]{0.62\textwidth}
        \centering
        \resizebox{\textwidth}{!}{%
        \begin{tabular}{l|ccc|ccc|ccc}
            \toprule
            & \multicolumn{3}{c|}{\textbf{Component}} & \multicolumn{3}{c|}{\textbf{Harvard-Glaucoma}} & \multicolumn{3}{c}{\textbf{CheXpert Plus}} \\
            \textbf{Configuration} & IFE & Within & Cross & AUC & ACC & F1 & AUC & ACC & F1 \\
            \midrule
            \proj (Full) & \checkmark & \checkmark & \checkmark & \textbf{0.850} & \textbf{0.768} & \textbf{0.786} & \textbf{0.966} & \textbf{0.919} & \textbf{0.920} \\
            w/o Within & \checkmark & & \checkmark & 0.838 & 0.748 & 0.779 & 0.964 & 0.913 & 0.913 \\
            w/o Cross & \checkmark & \checkmark & & 0.835 & 0.751 & 0.774 & 0.964 & 0.908 & 0.910 \\
            \bottomrule
        \end{tabular}%
        }
    \end{minipage}%
    \hfill
    \begin{minipage}[c]{0.35\textwidth}
        \centering
        \begin{normalsize}
        \begin{tabular}{lccc}
            \toprule
            \textbf{Input} & \textbf{MM LoRA} & \textbf{\proj} & $\Delta$ \\
            \midrule
            Both & 0.716 & 0.850 & +18.7\% \\
            w/o Image & 0.703 & 0.813 & +15.6\% \\
            w/o Text & 0.627 & 0.736 & +17.4\% \\
            \bottomrule
        \end{tabular}
        \end{normalsize}
    \end{minipage}
\end{table*}

\subsection{Main Results}
\label{sec:exp:main}

\Tbl{tbl:main_results} presents the main results on both datasets.
\proj consistently outperforms all baselines across both datasets and all metrics.

Multimodal LoRA performs poorly---even below single-modality baselines---due to shortcut learning: on Harvard-Glaucoma it achieves 0.716 AUC with high recall (0.904) but low precision (0.598), indicating the model predicts most samples as positive rather than learning from both modalities.
OGM-GE~\cite{peng2022balanced} and G-Blend~\cite{wang2020makes}, two gradient-based modality balancing methods, substantially improve AUC (0.835 and 0.837 on Harvard-Glaucoma), but exhibit a severe precision--recall imbalance---G-Blend reaches 0.857 precision yet only 0.572 recall.
On CheXpert Plus, both still fall short of the Text-only baseline (0.918/0.919 vs.\ 0.961 AUC), suggesting gradient-level balancing is insufficient when one modality dominates.

\proj outperforms all baselines on both datasets.
On Harvard-Glaucoma, it achieves the best AUC (0.850), accuracy (0.768), and F1 (0.786), surpassing OGM-GE/G-Blend by 1.6--1.8\% in AUC and 8.0--14.6\% in F1, while maintaining a balanced precision--recall profile (0.745/0.831).
On CheXpert Plus, \proj achieves 0.966 AUC, outperforming OGM-GE and G-Blend by over 5\%, and surpassing even the Text-only baseline (0.961)---demonstrating genuine multi-modal complementarity.
These results confirm that representation-level alignment is more effective than gradient-level balancing for multi-modal medical diagnosis with VLMs.

\begin{table}[!t]
    \centering
    \caption{Extension to \textbf{5-class multi-label} classification on CheXpert Plus (same 7K/1K/2K split; uncertain labels mapped to negative). \proj requires no architectural change. Mean over 3 seeds; largest std is 0.015 (mF1). Best in \textbf{bold}.}
    \label{tbl:multilabel}
    \vspace{-1em}
    \setlength{\tabcolsep}{3pt}
    \resizebox{\columnwidth}{!}{%
    \begin{tabular}{l|ccccc|ccc}
    \toprule
    & \multicolumn{5}{c|}{\textbf{Per-class AUC}} & \multicolumn{3}{c}{\textbf{Macro average}} \\
    \textbf{Method} & Card. & Edema & Cons. & Atel. & Effus. & mAUC & mAP & mF1 \\
    \midrule
    CGGM & 0.627 & 0.753 & 0.619 & 0.596 & 0.724 & 0.664 & 0.340 & 0.306 \\
    \proj & \textbf{0.819} & \textbf{0.791} & \textbf{0.703} & \textbf{0.679} & \textbf{0.810} & \textbf{0.760} & \textbf{0.473} & \textbf{0.480} \\
    \bottomrule
    \end{tabular}%
    }
\end{table}

\paragraph{Generalization to multi-label diagnosis.}
To test whether \proj is tied to binary tasks, we re-train it on CheXpert Plus as a 5-class multi-label problem (Cardiomegaly, Edema, Consolidation, Atelectasis, Pleural Effusion), and compare against CGGM~\cite{guo2024cggm}, a more recent classifier-guided gradient-modulation method.
The architecture is unchanged; only the objective is adapted in the standard way for multi-label targets, with the classification heads emitting five logits under a binary cross-entropy loss and $\Lwithin$ applied per label and averaged.
\proj reaches 0.760 mAUC, 0.473 mAP and 0.480 mF1 against 0.664, 0.340 and 0.306 for CGGM, a gain of $+0.097$ mAUC and $+0.174$ mF1 (\Tbl{tbl:multilabel}).
The improvement holds for all five findings, from $+0.039$ AUC on Edema to $+0.192$ on Cardiomegaly, indicating that the gains are not an artifact of the binary formulation.

\subsection{How Much Does \proj Reduce the Text Shortcut?}
\label{sec:exp:shortcut}

\begin{figure}[!t]
    \centering
    \includegraphics[width=\columnwidth]{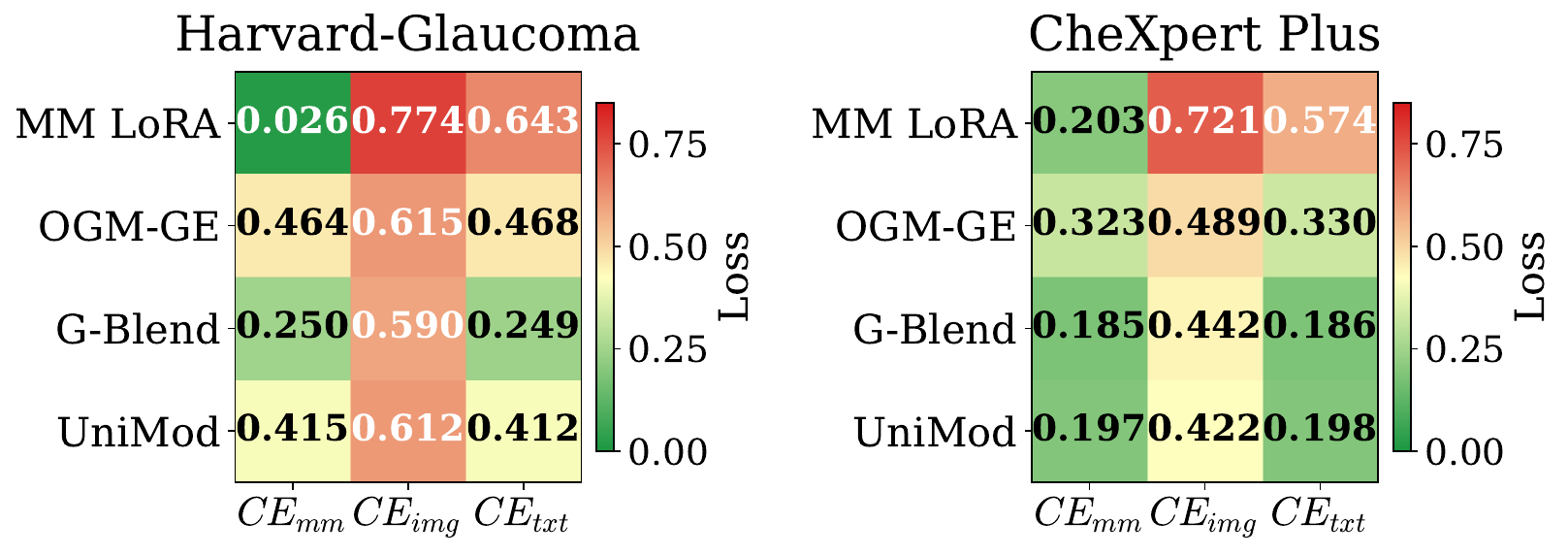}
    \vspace{-1em}
    \caption{Classification loss components at the end of training (greener is better optimized). Multimodal LoRA satisfies the fused objective ($CE_{mm}$=0.026) with $CE_{img}$=0.774 still high; \proj keeps all three terms balanced on both datasets.}
    \Description{Color-coded heatmap of the three classification loss components measured at the end of training, with training methods along one axis and the loss terms image-only cross-entropy, text-only cross-entropy, and multi-modal cross-entropy along the other, shown for both datasets. Green cells denote low loss and better optimization, while warmer cells denote high loss. The Multimodal LoRA row is strongly mixed: its multi-modal cross-entropy cell is near zero while its image-only cross-entropy cell is high, indicating the image encoder never learns. The UniMod row is uniformly green across all three components on both datasets, indicating balanced optimization.}
    \label{fig:loss_heatmap}
\end{figure}

The shortcut is directly measurable. Trained alone, the text branch reaches 0.814 AUC on Harvard-Glaucoma against 0.734 for the image branch, and 0.961 against 0.909 on CheXpert Plus (\Tbl{tbl:main_results}).
Text is the cheaper predictor, so a fused model can minimize its training objective by riding the text branch, and \Fig{fig:loss_heatmap} shows it doing so: Multimodal LoRA drives $CE_{mm}$ to 0.026 while $CE_{img}$ remains at 0.774, \ie, the fused loss is satisfied \emph{without} the image encoder becoming diagnostic.
This is a mechanism, not a score; independent unimodal supervision is designed to close that route.

\Tbl{tbl:robustness} compares the robustness of Multimodal LoRA and \proj when a modality is removed at test time.
Multimodal LoRA suffers significant performance degradation when either modality is missing, indicating over-reliance on cross-modal shortcuts.
In contrast, \proj maintains much higher performance with missing modalities, confirming that each modality has learned meaningful diagnostic features independently.

\begin{figure}[!t]
    \centering
    \includegraphics[width=\columnwidth]{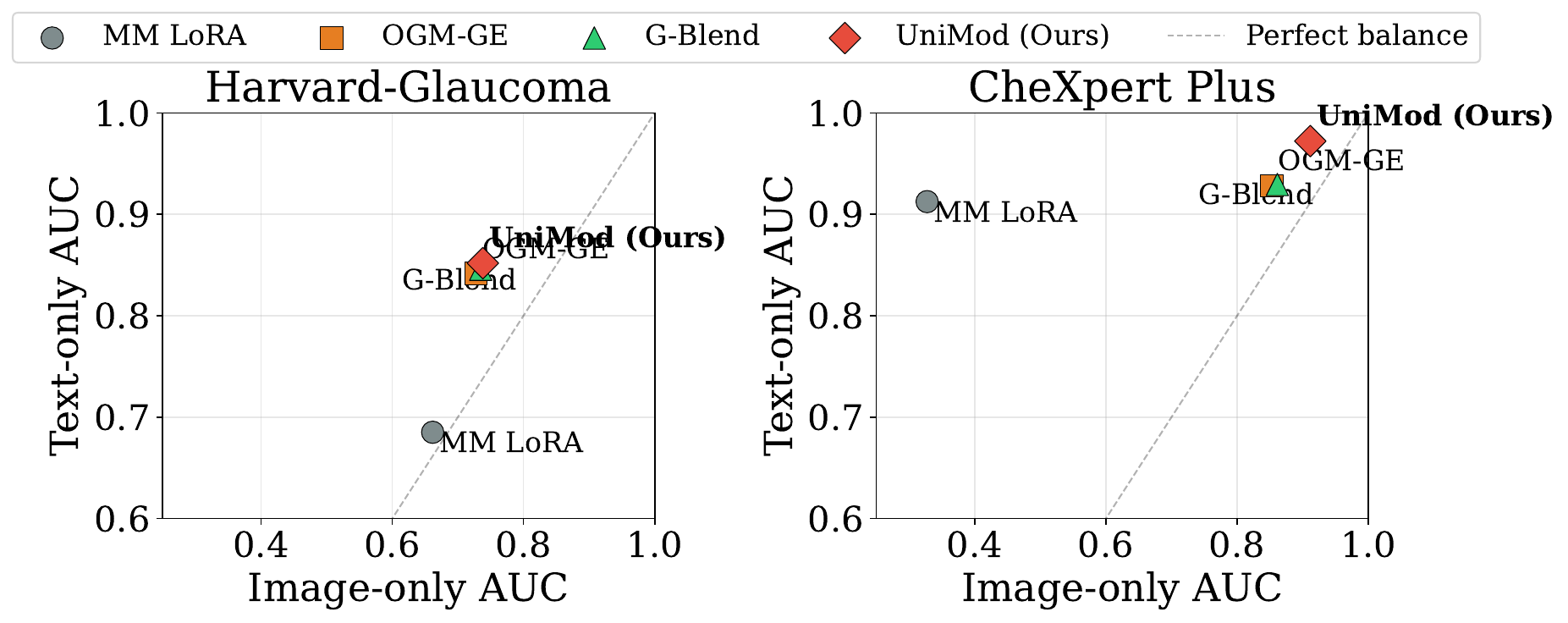}
    \caption{Single-modality AUC on Harvard-Glaucoma. Distance from the diagonal indicates modality imbalance; distance from the upper-right corner, how little either branch learned.}
    \Description{Scatter plot comparing single-modality performance on the Harvard-Glaucoma dataset. The horizontal axis is image-only AUC and the vertical axis is text-only AUC, with a diagonal reference line marking equal performance in both modalities. Each point is one training method. Methods lying near the upper-right corner achieve high AUC in both modalities, and methods lying close to the diagonal are balanced between them. UniMod sits closer to both the upper-right corner and the diagonal than the competing methods, which fall further from the diagonal because one modality dominates.}
    \label{fig:modality_balance}
\end{figure}

\Tbl{tbl:robustness} and \Fig{fig:modality_balance} measure the resulting reliance.
Removing either input at test time degrades Multimodal LoRA sharply, and its single-modality AUCs on Harvard-Glaucoma are low in both branches (image 0.662, text 0.685); \proj is higher and better balanced on both axes.
The asymmetry is starkest on CheXpert Plus (\Fig{fig:modality_gap}): Multimodal LoRA opens a 0.58 gap between text (0.91) and image (0.33) AUC, its image branch scoring below chance, whereas \proj reduces the gap to 0.06 at the highest absolute performance.
We note the scope of this evidence: these are input-ablation diagnostics of \emph{modality} reliance, not interventions on identified spurious features; they show which input the fused decision depends on, not which cues within an input are spurious.

\begin{figure*}[t]
    \centering
    \includegraphics[width=\textwidth,trim=0 300 0 0,clip]{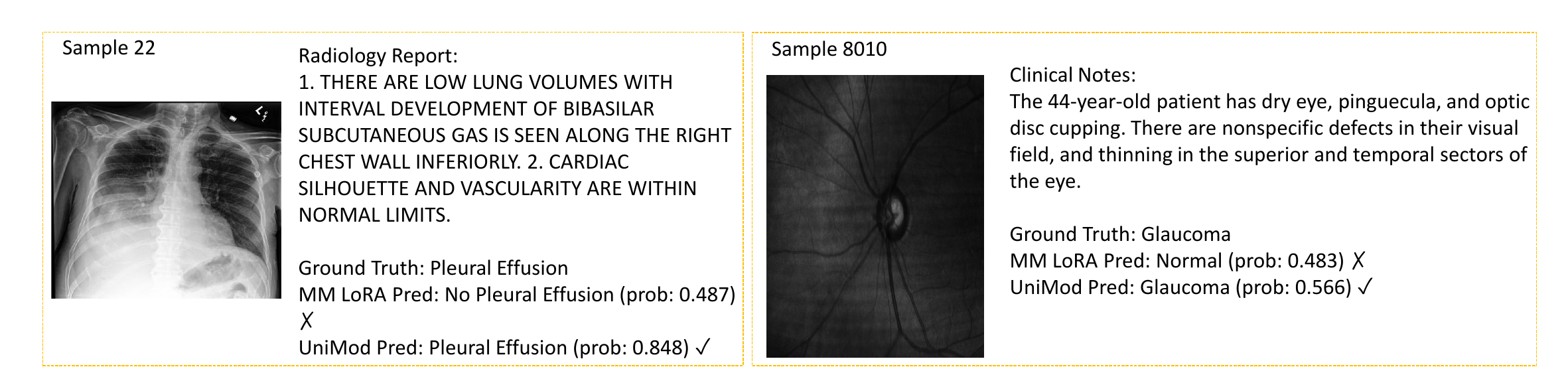}
    \caption{Case studies where \proj correctly classifies but Multimodal LoRA fails. Left: CheXpert Plus (chest X-rays with radiology reports for pleural effusion detection). Right: Harvard-Glaucoma (fundus images with clinical notes for glaucoma detection). \proj leverages visual features learned through independent feature extraction when clinical text is ambiguous. Chest radiographs \textcopyright{} Stanford AIMI; fundus images \textcopyright{} Harvard Ophthalmology AI Lab.}
    \Description{Two side-by-side case studies of individual patients that UniMod classifies correctly and Multimodal LoRA classifies incorrectly. The left panel is a CheXpert Plus case: a chest X-ray is shown together with an excerpt of its radiology report for pleural effusion detection, and the report wording is ambiguous. The right panel is a Harvard-Glaucoma case: a retinal fundus photograph is shown together with an excerpt of its clinical note for glaucoma detection, again with ambiguous text. In both panels the predictions of the two methods are listed against the ground-truth label, showing that when the clinical text is uninformative UniMod still reaches the correct diagnosis by relying on the visual features obtained through independent feature extraction, whereas Multimodal LoRA follows the misleading text.}
    \label{fig:case_studies}
\end{figure*}

\subsection{Ablation Studies}
\label{sec:exp:ablation}

We conduct ablation studies to analyze the contribution of each component in \proj.
\Tbl{tbl:ablation} shows the results.

Independent Feature Extraction (\Sect{sec:method:independent}) provides the largest improvement by forcing each modality to learn diagnostic features independently.
Cross-modal and within-modal alignment (\Sect{sec:method:cross}, \Sect{sec:method:within}) further improve performance by enabling knowledge transfer and structuring the representation space.
Removing either alignment component leads to performance drops on both datasets, with cross-modality alignment showing a slightly larger impact on Harvard-Glaucoma while both components contribute similarly on CheXpert Plus.
GradNorm (\Sect{sec:method:objective}) provides additional gains by automatically balancing the loss weights.
All configurations use PEFT via LoRA (\Sect{sec:method:lora}).

\paragraph{Loss component analysis.}
Under \proj all three terms stay balanced on Harvard-Glaucoma ($CE_{mm}$=0.415, $CE_{img}$=0.612, $CE_{txt}$=0.412; \Fig{fig:loss_heatmap}), \ie, each modality actively minimizes its own loss rather than deferring to the fused head; the Multimodal LoRA counter-example is analyzed in \Sect{sec:exp:shortcut}.

\begin{figure}[!t]

    \centering
    \includegraphics[width=.9\columnwidth]{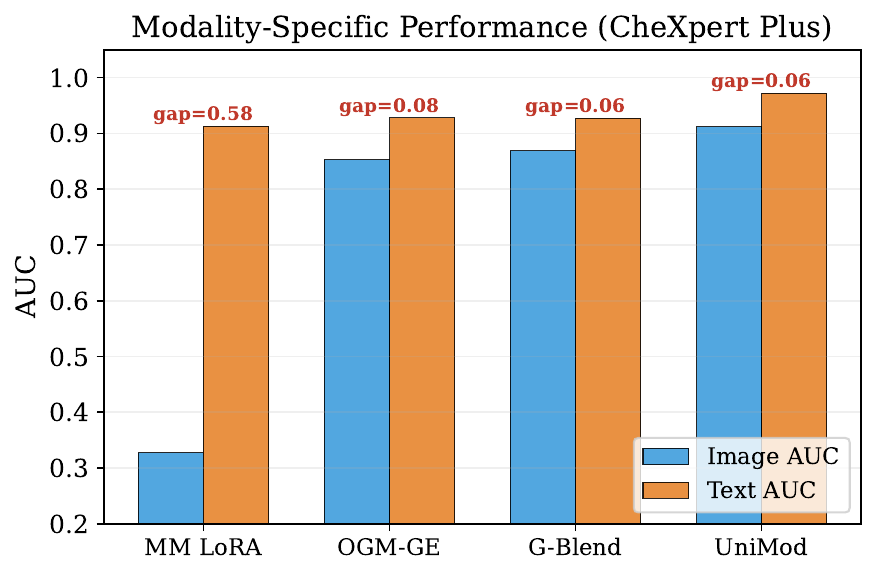}
    \caption{Modality-specific AUC on CheXpert Plus; annotations give the absolute image--text gap. The 0.58 gap of Multimodal LoRA is the size of its text shortcut---image AUC 0.33 is below chance---while \proj closes it to 0.06.}
    \Description{Grouped bar chart of modality-specific AUC on CheXpert Plus. For each training method there is one bar for image-only AUC and one bar for text-only AUC, and each pair is annotated with the absolute gap between them. Multimodal LoRA shows the most severe imbalance, with a gap of 0.58 between a high text bar and a much lower image bar. UniMod shows the smallest gap between its two bars while also reaching the highest overall performance.}
    \label{fig:modality_gap}
\end{figure}

\begin{figure}[t]
    \centering
    \vspace*{10pt}
    \includegraphics[width=\columnwidth]{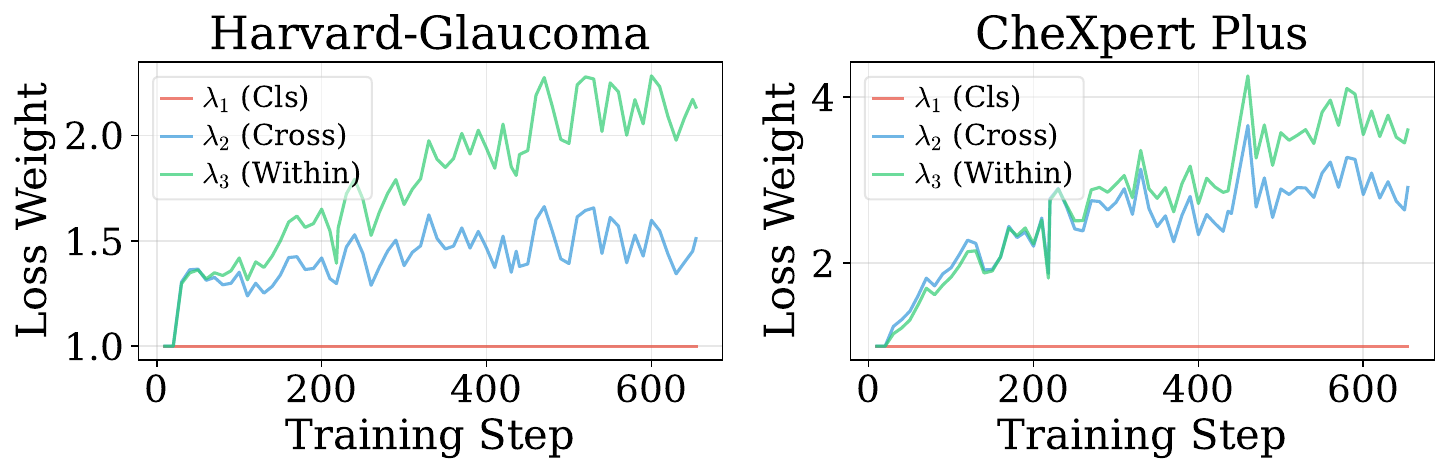}
    \caption{GradNorm weight evolution. $\lambda_1$ is fixed at 1.0; $\lambda_2$ (cross-modal) and $\lambda_3$ (within-modal) are adjusted automatically, rising up to fourfold on CheXpert Plus, consistent with its greater modality imbalance.}
    \Description{Line plot of the GradNorm loss weights over the course of training, with training step on the horizontal axis and weight value on the vertical axis, drawn for both datasets. The classification weight lambda-1 is a flat horizontal line held at 1.0. The cross-modality weight lambda-2 and the within-modality weight lambda-3 start near their initial values and are then adjusted automatically as training proceeds. Both alignment weights settle at higher values on CheXpert Plus than on Harvard-Glaucoma, reflecting the greater modality imbalance of the CheXpert Plus data.}
    \label{fig:gradnorm}
\end{figure}

\paragraph{Dynamic loss balancing.}
On both datasets GradNorm raises $\lambda_2$ (cross-modal) and $\lambda_3$ (within-modal) relative to the fixed $\lambda_1$ (classification; \Fig{fig:gradnorm}), indicating that the alignment losses need larger weights to balance the classification objective.

\paragraph{Why gradient modulation is insufficient.}
OGM-GE~\cite{peng2022balanced} and G-Blend~\cite{wang2020makes} adjust gradients or loss weights but do not change the training objective: only the fused prediction loss is supervised.
As shown in \Fig{fig:loss_heatmap}, $CE_{img}$ remains high even with gradient modulation, because suppressing the dominant modality does not guarantee the weaker modality actively learns.
\proj instead provides explicit unimodal supervision and structures the representation space via cross-modal and within-modal alignment~\cite{wang2020understanding}, achieving the best performance and modality balance (\Fig{fig:modality_gap}).





\subsection{Case Study}
\label{sec:exp:case}
\Fig{fig:case_studies} shows one case per dataset in which \proj succeeds and Multimodal LoRA fails.
In the CheXpert Plus case the report describes indirect findings (\eg, ``low lung volumes,'' ``bibasilar subcutaneous gas'') without naming pleural effusion; Multimodal LoRA returns 0.487, while \proj detects the effusion at 0.848.
In the Harvard-Glaucoma case the note is suggestive but inconclusive (\eg, ``optic disc cupping,'' ``nonspecific defects''), and Multimodal LoRA again lands near chance at 0.483 against \proj's 0.566.
In both, Multimodal LoRA's prediction sits close to 0.5: absent an explicit diagnostic keyword it effectively abstains rather than reading the image, which is the behavior \Fig{fig:loss_heatmap} shows in the objective itself.
\proj instead commits, because the image encoder was required to be diagnostic on its own.

\section{Conclusion}
\label{sec:conclusion}

We presented \proj, a framework that mitigates shortcut learning in multi-modal medical diagnosis by requiring each modality to predict the diagnosis on its own, combining independent unimodal supervision with cross-modality and within-modality alignment under GradNorm weighting.
On Harvard-Glaucoma and CheXpert Plus it improves over standard multi-modal fine-tuning and over modality-balancing baselines, and it transfers to 5-class multi-label diagnosis without architectural change.
Our analysis indicates that gradient modulation alone does not remove the shortcut: explicit unimodal supervision is what drives each modality to develop diagnostic capability.
The components are modality- and task-agnostic, but the evidence here comes from two single-institution datasets. Cross-institutional evaluation, further modalities (CT, MRI, physiological signals~\cite{johnson2019mimic}), dense prediction tasks~\cite{ronneberger2015u} and integration with clinical decision support~\cite{singhal2023large,liu2019comparison} remain future work.

\newpage
\begin{acks}
This research used data provided by the Stanford Center for Artificial Intelligence in Medicine and Imaging (AIMI). AIMI curated a publicly available imaging data repository containing clinical imaging and data from Stanford Health Care, the Stanford Children's Hospital, the University Healthcare Alliance and Packard Children's Health Alliance clinics provisioned for research use by the Stanford Medicine Research Data Repository (STARR). Fundus images are reproduced from the Harvard-Glaucoma dataset \copyright{} Harvard Ophthalmology AI Lab, licensed CC BY-NC-ND 4.0. All images are used for non-commercial research purposes.
\end{acks}

\balance
\bibliographystyle{ACM-Reference-Format}
\bibliography{references}

\end{document}